\documentclass{article}
\usepackage{spconf,amsmath,graphicx}

\title{Siamese Networks with Location Prior\\ for Landmark Tracking in Liver Ultrasound Sequences}

\name{Alvaro Gomariz, Weiye Li, Ece Ozkan, Christine Tanner, Orcun Goksel\thanks{Funding provided by Swiss National Science Foundation.}}
\address{Computer-assisted Applications in Medicine, ETH Zurich, Switzerland}
\begin{document}
%
\maketitle
\begin{abstract} 
Image-guided radiation therapy can benefit from accurate motion tracking by ultrasound imaging, in order to minimize treatment margins and radiate moving anatomical targets, e.g., due to breathing. 
One way to formulate this tracking problem is the automatic localization of given tracked anatomical landmarks throughout a temporal ultrasound sequence.
For this, we herein propose a fully-convolutional Siamese network that learns the similarity between pairs of image regions containing the same landmark.
Accordingly, it learns to localize and thus track arbitrary image features, not only predefined anatomical structures. 
We employ a temporal consistency model as a location prior, which we combine with the network-predicted location probability map to track a target iteratively in ultrasound sequences.
We applied this method on the dataset of the Challenge on Liver Ultrasound Tracking (CLUST) with competitive results, where our work is the first to effectively apply CNNs on this tracking problem, thanks to our temporal regularization.

\end{abstract}
%
%
\section{INTRODUCTION}
\label{sec:intro}
Ultrasound (US) imaging provides excellent temporal resolution and is a non-invasive modality, which together make it ideal for image-guidance of procedures, including of radiation therapy (RT).  RT requires precise and conformal application of radiation dose in space, for which organ motion due to internal body movements (e.g. breathing) is a major challenge~\cite{intro_motion}. 
Tracking treatment target location in US is a promising approach, given that real-time and accurate tracking algorithms can be developed.
Image-based tracking techniques proposed in the literature include block matching \cite{Shepard17}, optical-flow \cite{Makhinya15, Williamson2018}, and supporter models~\cite{Ozkan2017}. Nevertheless, these methods are either slow or require often substantial parameter tuning to optimize for a particular image and landmark appearance.  
We propose herein a solution based on Convolutional Neural Networks (CNNs), which are fast at inference time, and their adaption to new data distributions is often straightforward given annotations. 
CNNs perform particularly well on classification tasks, but tracking has remained more of a challenge with CNNs so far. 
To the best of our knowledge, CNNs have only been applied to the ultrasound tracking problem in~\cite{nouri2015liver} as a metric learning framework that aims to minimize the distance between patches containing the same landmark at the center. 
However, this model is not fully convolutional and is therefore relatively slow. 
Furthermore, this approach fails to effectively account for any temporal information, which is crucial when similar or repetitive structures exist such as many vessels in the liver.

Recently, fully-convolutional Siamese (SiameseFC) networks for similarity learning have been applied successfully on tracking problems for natural scenes from camera images~\cite{Valmadre_2017_CVPR, bertinetto2016fully}. 
These methods aim to learn the similarity between a \emph{template} image that contains a specific object of interest and a \emph{search} image where a similar looking object is to be found. 
For this purpose, two identical CNNs are trained with their respective template and search images to represent arbitrary objects in an embedding, which can be used for effective comparison. 
Cross-correlation is applied to produce a similarity score map, from which the maximum value is chosen as the predicted landmark location.

This original SiameseFC~\cite{bertinetto2016fully} aims to detect a specific object or objects in a given image. We propose herein to adapt this method for finding targeted anatomical locations in consecutive frames. We achieve this by learning the image similarity between corresponding target locations via a customized ground truth representation and loss definition. 
To promote temporal consistency, we augment the similarity maps with a location prior based on the entire preceding tracked path.

\section{METHODS}
\label{sec:methods}

\begin{figure}[t]
\begin{minipage}[b]{1.0\linewidth}
  \centering
  \centerline{\includegraphics[width=8.5cm]{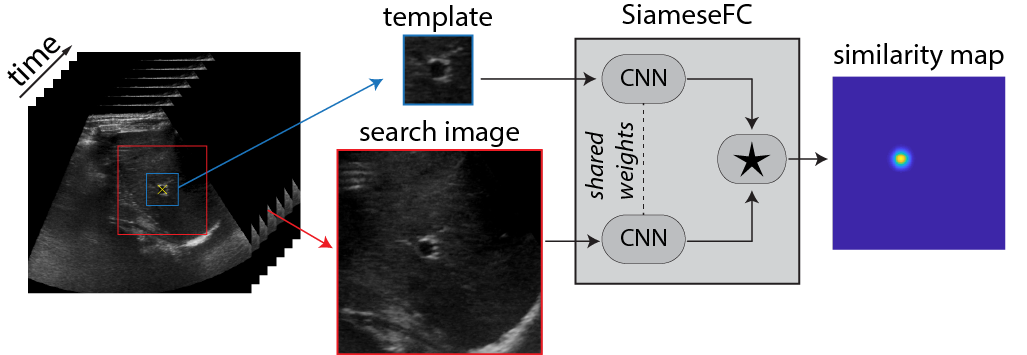}}
\end{minipage}
\caption{Our adaptation of SiameseFC for tracking in US sequences. A template-to-image similarity map is learned as the cross-correlation of the result from the Siamese CNNs.}
\label{fig:siamesefc}
\end{figure}

\vspace{.5ex}\noindent{\bf SiameseFC for similarity learning.}
Our SiameseFC was adapted from the method described in~\cite{Valmadre_2017_CVPR} for learning inter-frame similarity from annotated landmark locations as illustrated in Fig.\ref{fig:siamesefc}. It applies an identical CNN $f$ on both the template image $p$ (which contains the object of interest) and the search image $q$ to extract representative embeddings that can be effectively compared. This comparison is implemented as a cross-correlation layer (represented by the $\star$ operator) between the sliding window $f(p)$ and the search region $f(q)$, which results in the similarity function $S$
\begin{equation}
\label{eq:siameseFC}
S(p,q)=f(p)\star f(q)
\end{equation}
We defined the template $p$ as the landmark region in the first (annotated) frame.
A sufficiently large search region around this point in all subsequent frames are then used as $q$ search image. 
As $f$, we used the convolutional stage of the AlexNet architecture~\cite{AlexNet}, with batch normalization and ReLU activation after each convolutional layer except the very last one. 
In the original SiameseFC work~\cite{Valmadre_2017_CVPR}, the CNN employs a pixel-wise logistic loss on the similarity map. 
Its ground truth is generated by setting pixels within a radius from the landmark to +1, and elsewhere to -1.  
We first attempted this approach in its native form, weighting each pixel by its class cardinality or weighting with a Gaussian function centered at the manual annotation. 
Across several variations initially, only one particular combination resulted in successful results, where we
\vspace{-1.5em}\begin{itemize}  \setlength{\parskip}{0pt} \setlength{\itemsep}{0pt plus 1pt}
    \item generate the ground truth as a 2D \emph{Gaussian map} 
$G_{\mu,\sigma}(u)=\frac{1}{\sigma \sqrt {2\pi } }e^{ -\left( {u - \mu } \right)^2/2\sigma ^2}$
centered at the ground truth landmark location $y$, and
    \item use an L2-loss to compare this with the tracking output
\begin{equation}
\label{eq:loss}
L(S, y) = \frac{1}{2}\sum_{u \in S}\big(S(u) - G_{y,\sigma}(u)\big)^2
\end{equation}
where $u$ corresponds to individual pixel locations in $S$. 
\end{itemize}
\vspace{-\topsep}
At the training stage we randomly pair annotated images from the same sequence as templates and search images showing the same annotated landmark at different time instances. 
For the prediction stage, the template is taken from the annotation at the first frame, and the search images are taken as all subsequent frames in the sequence. 

\begin{figure}[t]
\begin{minipage}[b]{1.0\linewidth}
  \centering
  \centerline{\includegraphics[width=8.5cm]{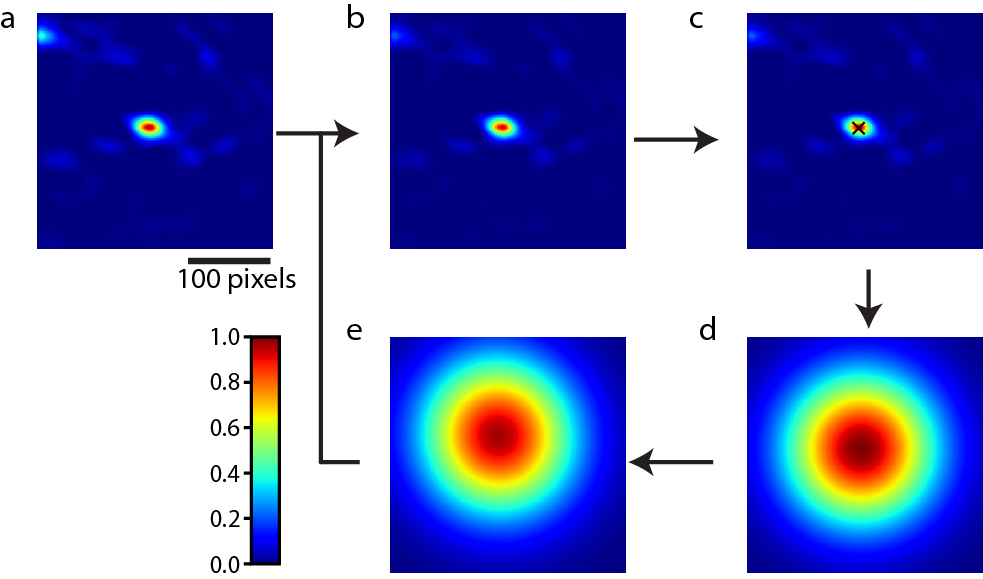}}
\end{minipage}
\caption{Location prior using previous predictions. (a) the similarity map $S_t$ from SiameseFC, which is weighted by $R_t$ in Eq.(\ref{eq:motion}) to produce $S'_t$ in (b). Given Eq.(\ref{eq:pred_lm}) a landmark $x_i$ is predicted at the maximum as in (c), from which a 2D Gaussian map $G_{x_t,\sigma}$ is generated as in (d), used then to update a running average as the location prior as shown in (e). }
\label{fig:motion}
\end{figure}

\vspace{.5ex}\noindent{\bf Temporal consistency model for landmark tracking.}
\label{sec:methods_motion}
With the above approach, a similarity metric to find a landmark in subsequent frames can be learned well. 
However, this incorporates no temporal information in its given naive form. 
In preliminary results, we saw this as a main limitation, especially when similar anatomical features come in proximity and the tracker switches to such false targets. 
We propose to augment the similarity maps with a location prior as illustrated in Fig. \ref{fig:motion}.
To that end, we build a temporal consistency model based on the history of all predicted landmark positions in the preceding frames.
This acts as a location prior and a confidence model of where to most likely expect a target location. 
This model somewhat regularizes the predicted similarity score $S$, helping by avoiding the prediction of landmarks in unlikely regions. 
At time $t$, we first update the location prior $G^*(u)$ at all positions $u$ as a running average, i.e.
\begin{equation}
\label{eq:regularizer}
G^* \leftarrow G^* + \frac{G_{x_i,\sigma}-G^*}{|t|}
\end{equation}
A temporal regularizer $R_t = 1 - G^*$ is then used to weight the similarity map to update it with temporal consistency as
\begin{equation}
\label{eq:motion}
S_t'(u) = S_t(u)\big(1 - w_tR_t(u)\big) 
\end{equation}
where the parameter $w_t$ determines the weighting of the location prior at time $t$.
To avoid new positions being penalized during early tracking iteration when the model is being first constructed, this weight is set to increase with time as follows:
\begin{equation}
\label{eq:cert_motion}
w_t = k\tanh(t,\tau) 
\end{equation}
where constant $k$ balances the maximum contribution of $w_t$ in $S_t$ and $\tau$ is a constant defining how fast $w_t$ should grow.
We define this growth rate empirically as approximately the time for one breathing cycle. 

Given anatomical constraints, the landmark is assumed that it cannot move further than a predefined distance of $d_{\max}$ between two frames.
Accordingly, the maximum of $S_t'$ within a radius of $d_{\max}$ from the previous location is chosen as the landmark location $x_t$, i.e.
\begin{equation}
\label{eq:pred_lm}
    x_t  = \arg\max_{u}S'(u) \quad \textit{s.t.}  \quad \lVert x_t-x_{t-1} \rVert < d_{\max} 
\end{equation}

\vspace{.5ex}\noindent{\bf Implementation}
was achieved with TensorFlow~\cite{tensorflow2015-whitepaper}, with the network training and the experiments ran on an Nvidia GeForce GTX TITAN X GPU.
Based on initial empirical tests, we employed a batch size of 16 images, a learning rate of $1e$-$7$, and trained for 100 epochs with the Adam optimizer. 
We set $\sigma$$=$$2.16$\,mm in Eq.(\ref{eq:loss}), $\sigma$$=$$17.28$\,mm in Eq.(\ref{eq:regularizer}), $k=0.5$ and $\tau$$=$$50$ in Eq.(\ref{eq:cert_motion}), and $d_{\max}$$=$$8.64$\,mm in Eq.(\ref{eq:pred_lm}).

\section{RESULTS AND DISCUSSION}
\label{sec:results}
\noindent{\bf Dataset.}
We applied our method to 2D liver US sequences provided by the Challenge on Liver Ultrasound Tracking (CLUST) \cite{de2018evaluation}, which was prepared to address the localization of anatomical landmarks under respiratory motion in the liver. The dataset contains 2D liver US sequences from four different clinical centers, with durations ranging from 60 to 330 seconds, at varying spatial and temporal resolutions. 
Each sequence has one or more landmarks annotated in the first frame, which are the landmarks to be tracked for the remaining frames. 
24 sequences are provided by CLUST as the training set, from which we used 20 for training our CNN and the remaining 4 for validation. 
The test set consists of 40 sequences with a total of 85 landmarks for tracking annotated on the initial frames. The corresponding annotations for the remaining frames are inaccessible to participants and are evaluated by the organizers upon submission. Mean, standard deviation, and 95 percentile of the errors are reported, which are calculated as the Euclidean distance between the manual ground truth and the predicted landmark at each of the annotated frames.

\vspace{.5ex}\noindent{\bf Patches.} 
To normalize anatomical feature sizes, we resampled all the images to 0.27\,mm/pixel, which is the maximum resolution available in the dataset. 
The template image for each sequence was created by cropping a 127x127 region around the initial annotation in the first frame of the sequence. 
This region was considered to contain the spatial context required for tracking.
Search images were cropped in all subsequent frames as a 407x407 region around the position of the initial landmark. 
This size includes a margin for the maximum liver motion possible due to anatomical constraints. 

\vspace{.5ex}\noindent{\bf Loss.} 
In preliminary tests on our validation set of 4 sequences (with 2 of them containing 2 annotations to track, thus, 6 landmarks in total), the original SiameseFC design~\cite{Valmadre_2017_CVPR} with its binary ground truths and logistic loss function was not converging to yield viable similarity maps. 
Distant pixels were being activated in results, suggesting the model being incapable of accurately discriminating the under-represented true-positive locations from the false-negative ones. 
Weighting the loss function by the class cardinality to avoid class imbalance did not solve the problem. 
Logistic loss was found not optimal for the US images in which false-negative regions could easily have very similar features as the true-positive ones; contrary to natural scenes used in the original SiameseFC.
To address this, we employed L2-loss to a probabilistic ground truth as a 2D Gaussian function centered at the desired landmark for smooth and derivable boundaries between classes.  
This resulted in substantially lower scores on our validation set, confirming the learning of a similarity map for landmarks. 

\begin{figure}[t]
\begin{minipage}[b]{0.49\linewidth}
  \centering
  \centerline{\includegraphics[width=4.1cm]{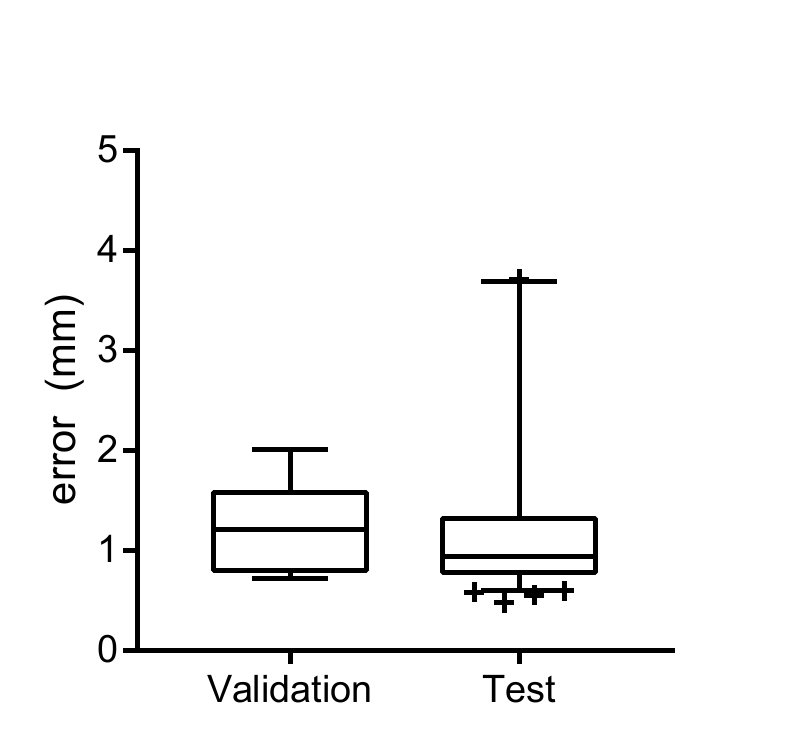}}
  \centerline{(a) Mean error}
\end{minipage}
\hfill
\begin{minipage}[b]{0.49\linewidth}
  \centering
  \centerline{\includegraphics[width=4.1cm]{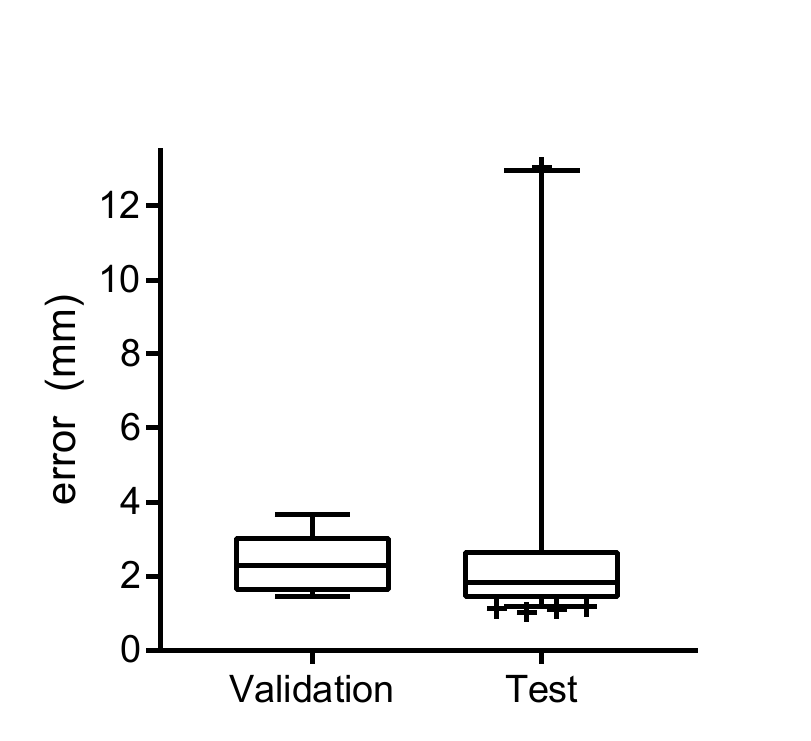}}
  \centerline{(b) 95\%ile error}
\end{minipage}
\caption{Distribution of errors for the validation set (n=6) and the test set (n=85). For the test set, 3 points fall out of the graph area for both the mean and the 95\%ile error.}
\label{fig:res_metrics}
\end{figure}

\begin{figure}
\begin{minipage}[b]{1.0\linewidth}
  \centering
  \centerline{\includegraphics[width=8.5cm]{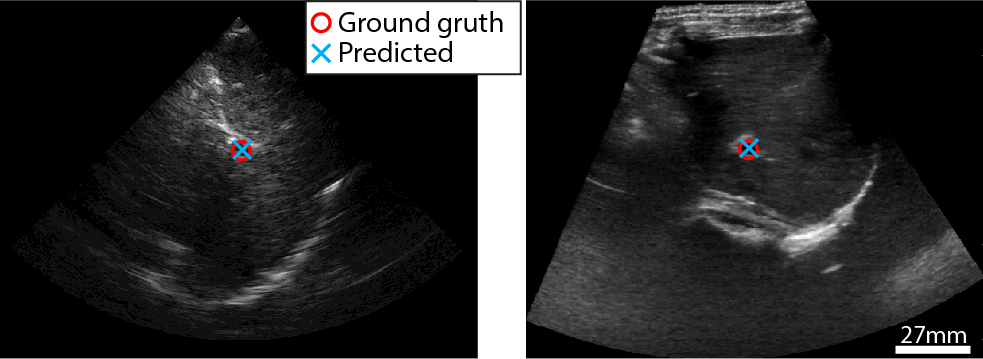}}
\end{minipage}
\caption{Visualization of ground truth and predicted landmarks in randomly selected frames from two different sequences in the validation set.}
\label{fig:res_im}
\end{figure}

\vspace{.5ex}\noindent{\bf Temporal Consistency Model.}
By incorporating our location prior for tracking, we obtained an error of 1.29$\pm$0.74\,mm, cf. Fig.\,\ref{fig:res_metrics}(a), with a very close tracking to the ground truth as seen in Fig.\ref{fig:res_im}. 

\begin{figure}[t]
\begin{minipage}[b]{1.0\linewidth}
  \centering
  \centerline{\includegraphics[width=8.5cm]{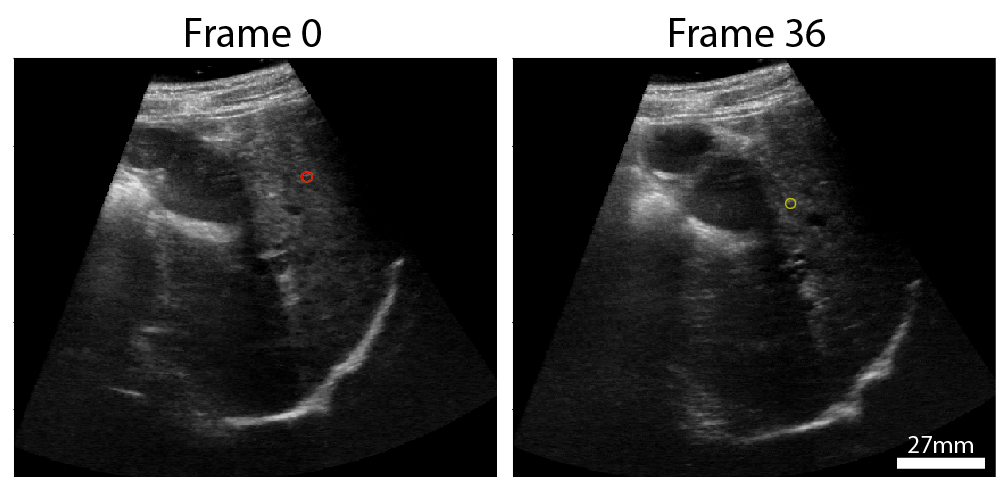}}
\end{minipage}
\caption{Sample frames from the test sequence with the highest mean error, showing the annotation provided in frame 0 (left) and the tracked location in frame 36 (right). Although no ground truth is provided, it can be seen that the predicted landmark marks a different structure in frame 36. The original structure seems to have changed substantially, rendering not even distinguishable for a human observer.}
\label{fig:rest_imtest}
\end{figure}

\vspace{.5ex}\noindent{\bf Test set.}
We submitted our method for evaluation on the test set of the open CLUST challenge~\cite{de2018evaluation}, in which we obtained an error of 1.34$\pm$2.57\,mm. 
Despite overall high accuracy, for a few test sequences the errors are quite large; see Fig. \ref{fig:res_metrics}(b). 
A visual inspection reveals that these occur when there are very similar features in proximity and the tracked location abruptly switches to the false one (see the example in Fig.\,\ref{fig:rest_imtest}). 
We could not find any frame on the validation set with this source of error. We believe that this error is due to the lack of motion features on the CNN part of our method.
\begin{table}[t]
\begin{tabular}{|l|rrr|}
\hline
Method                   & \multicolumn{1}{c}{Mean} & \multicolumn{1}{c}{$\sigma$} & \multicolumn{1}{c|}{95\%ile} \\ \hline
Shepard A., et al.      
& 0.72 & 1.25                  & 1.71    \\ \hline
Williamson T., et al.    
& 0.74 & 1.03                  & 1.85    \\ \hline
(anonymous)                
& 1.11 & 0.91                  & 2.68    \\ \hline
Hallack A., et al. 
& 1.21 & 3.17                  & 2.82    \\ \hline
\textbf{SiameseFC + regularization}
& 1.34 & 2.57                  & 2.95    \\ \hline
Makhinya M. \& Goksel O. 
& 1.44 & 2.80                  & 3.62    \\ \hline
Ihle F. A.               
& 2.48 & 5.09                  & 15.13   \\ \hline
Kondo S.                 
& 2.91 & 10.52                 & 5.18    \\ \hline
Nouri D. \& Rothberg A. 
& 3.35 & 5.21                  & 14.19   \\ \hline
\end{tabular}
\label{table:res_official}
\caption{Test-set results of the open CLUST challenge (extracted from 
clust.ethz.ch/results.html) in mm, ranked according to increasing mean error. Our method is displayed in bold.}
\end{table}

\vspace{.5ex}\noindent{\bf Discussion.}
In contrast to conventional SiameseFC, our proposed Gaussian soft ground truth with L2-loss is able to learn US tracking problem despite similar looking false-positive alternatives.
Our regularization effectively penalizes misleading similarities with a location prior, built based on a relatively simple form of temporal information. In contrast, many methods that perform superior to ours in CLUST use sophisticated motion models or priors, such as motion dynamics through Kalman filtering, which would also be possible to incorporate in our method in the future.
Alternatively or in addition, a long short-term memory (LSTM) unit can be incorporated in our approach to integrate similarity maps throughout sequences. 

\vspace{.5ex}\noindent{\bf Relevance.}
We show relevant error ranges and 95\% errors, compared to typical RT treatment margins of 5 to 10\,mm. 
Given that speed and adaptability to different datasets is of utmost importance for RT image guidance, we believe our CNN-based approach can provide an ideal solution, as CNNs can run quite fast, especially on GPU. 
Average inference time of our proposed method is 9.4\,ms per frame, much faster than the acquisition rate of the US sequences employed.

\section{CONCLUSION}
\label{sec:conclusion}
Landmark tracking in US sequences is a challenging and important topic given its relevance in clinical settings. While previous methods have achieved errors similar to humans, they are often slow and have difficulties to generalize to different data distributions. While CNNs can tackle these obstacles in other fields, their application to medical image tracking has been under-explored. 
We propose herein an adaptation of SiameseFC to accurately learn similarity maps from a landmark in US to a search region, which we augment with a location prior for temporal consistence. 
Given our contributions, competitive results have been achieved using fast, extendable CNNs.
Future directions include more sophisticated motion models and LSTMs for temporal consistency.


\bibliographystyle{IEEEbib}
\bibliography{refs}

\end{document}